\documentclass[lettersize,journal]{IEEEtran}
\usepackage{amsmath,amsfonts}
\usepackage{algorithmic}
\usepackage{algorithm}
\usepackage{array}
\usepackage[caption=false,font=normalsize,labelfont=sf,textfont=sf]{subfig}
\usepackage{textcomp}
\usepackage{stfloats}
\usepackage{url}
\usepackage{verbatim}
\usepackage{graphicx}
\usepackage{cite}
\usepackage{amsmath,bm}
\usepackage{color}
\usepackage{threeparttable}

\begin{document}


\title{Global-Regularized Neighborhood Regression for Efficient Zero-Shot Texture Anomaly Detection}

\author{Haiming Yao,
        Wei Luo,
        Yunkang Cao,
        Yiheng Zhang, \\
        Wenyong Yu,~\IEEEmembership{Senior Member,~IEEE,} 
        and Weiming Shen~\IEEEmembership{Fellow,~IEEE}
        
\thanks{Manuscript received XX XX, 20XX; revised XX XX, 20XX. This study was supported in part by the National Natural Science Foundation of China (Grant No. 52375494) (Corresponding author: Wenyong Yu.)}
\thanks{Haiming Yao, Wei Luo are with the State Key Laboratory of Precision Measurement Technology and Instruments, Department of Precision Instrument, Tsinghua University, Beijing 100084, China. (e-mails: $\{$yhm22,luow23$\}$@mails.tsinghua.edu.cn).}
\thanks{Yunkang Cao, Yiheng Zhang, Wenyong Yu, and Weiming Shen are with the State Key Laboratory of Digital Manufacturing Equipment and Technology, School of Mechanical Science and Engineering, Huazhong University of Science and Technology, Wuhan 430074, China(e-mails:cyk$\_$hust@hust.edu.cn, yihengzhang@hust.edu.cn, ywy@hust.edu.cn, wshen@ieee.org).}
}

\markboth{SUBMISSION TO IEEE Transactions on Systems, Man, and Cybernetics: Systems}%
{Shell \MakeLowercase{\textit{et al.}}: A Sample Article Using IEEEtran.cls for IEEE Journals}


\maketitle

\begin{abstract}

Texture surface anomaly detection finds widespread applications in industrial settings. However, existing methods often necessitate gathering numerous samples for model training. Moreover, they predominantly operate within a close-set detection framework, limiting their ability to identify anomalies beyond the training dataset. To tackle these challenges, this paper introduces a novel zero-shot texture anomaly detection method named Global-Regularized Neighborhood Regression (GRNR). Unlike conventional approaches, GRNR can detect anomalies on arbitrary textured surfaces without any training data or cost. Drawing from human visual cognition, GRNR derives two intrinsic prior supports directly from the test texture image: local neighborhood priors characterized by coherent similarities and global normality priors featuring typical normal patterns. The fundamental principle of GRNR involves utilizing the two extracted intrinsic support priors for self-reconstructive regression of the query sample. This process employs the transformation facilitated by local neighbor support while being regularized by global normality support, aiming to not only achieve visually consistent reconstruction results but also preserve normality properties. We validate the effectiveness of GRNR across various industrial scenarios using eight benchmark datasets, demonstrating its superior detection performance without the need for training data. Remarkably, our method is applicable for open-set texture defect detection and can even surpass existing vanilla approaches that require extensive training.

\end{abstract}


\section{Introduction}

\IEEEPARstart{T}{exture} anomaly detection aims to identify irregular patterns that deviate from normal contexts for given categories. These texture anomalies typically manifest as localized defects or damages. For reasons pertaining to product quality control \cite{r8}, safety assurance \cite{r50}, and other considerations, detecting such defects is an essential process in numerous industrial scenarios, like industrial inspection on product surfaces \cite{r8}, pavement surfaces \cite{r9}, and fabrics \cite{r10}.  Machine vision-based Automatic Optical Inspection (AOI) systems\cite{r59}  \cite{r11} have presented promising flexible and robust detection performance and garnered increasing attention for texture anomaly defect detection. 




From a holistic perspective on advancements, recent methodologies for texture anomaly detection can be classified into the following categories as shown in Fig. 1. Supervised methods \cite{r1,r3,r6,r60} (Fig. 1 (a)) have demonstrated robust performance across various domains, including strip defect detection \cite{r12}, fabric defect detection \cite{r10}, and pavement crack detection \cite{r13}. However, these supervised methods often rely on large anomalous defective datasets and corresponding annotations. For instance, expensive pixel-level annotations are required in supervised segmentation applications \cite{r12}. Obtaining such data and annotations demands significant effort and expertise. Moreover, this type of approach is a "defect close-set" solution, where the model is restricted to the defect types encountered during model training and lacks generalizability to novel defect types. 


In order to reduce the labor involved in collecting defective samples and annotating data, unsupervised-learning-based texture anomaly detection methods~\cite{r8,r11} shown in Fig. 1(b) were subsequently developed, in which only non-defective normal samples are required for training. This implies that it can detect unknown anomalous defects and has drawn widespread attention recently~\cite{r14,r16}. Nevertheless, a substantial quantity of normal samples remains essential for model training, also maintaining a “texture close-set” detection paradigm wherein the model can solely identify texture types present within the training dataset. In essence, the aforementioned two types of methods can be categorized as large-scale data-driven approaches. To investigate data-efficient approaches for texture anomaly defect detection, recent studies have extensively delved into the few-shot learning paradigm \cite{r17} \cite{r18} as shown in Fig. 1(c). This paradigm necessitates only a limited number of samples (typically fewer than 10) as prior information for model training. In comparison to the aforementioned methods, it offers a more flexible deployment approach. However, these few-shot methods are also within the texture close-set paradigm.

Modern texture anomaly detection systems require enhanced capabilities and functionalities. Primarily, there is a need for improved data and labor efficiency, aiming to achieve accurate detection while minimizing data and labor requirements. Both supervised and unsupervised methods discussed earlier rely on extensive datasets for deployments, which hinders their ability to meet these specified requirements. Additionally, an essential aspect is the open-set detection capability. In real-world industrial settings, encountering various unknown textured surfaces is common. In such scenarios, the inspection system lacks prior information about the surface, rendering the aforementioned methods ineffective.

\begin{figure*}[t]
\centerline{\includegraphics[width=180mm]{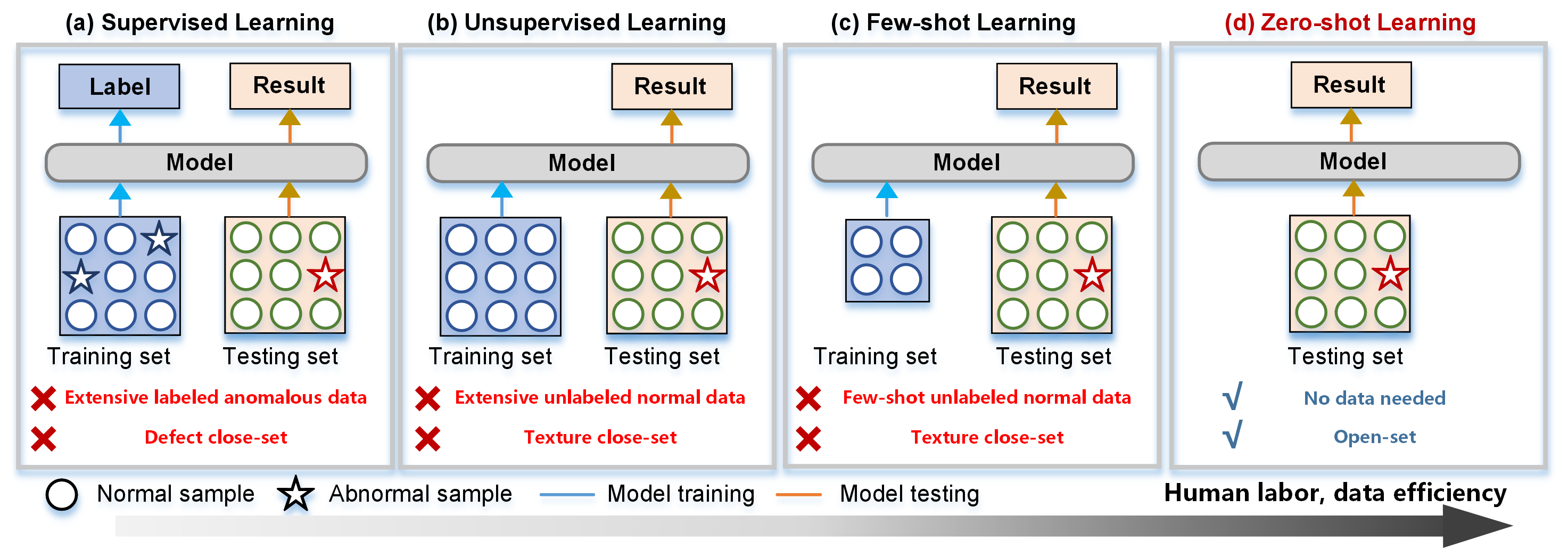}}
\caption[width=175mm]{
A comprehensive comparison of existing textured surface anomaly defect detection methods. (a) Supervised learning methods necessitate a substantial amount of labeled data for training and operate within a defect close-set detection paradigm. This means they can only identify defect types present in the training dataset. (b)-(c). Unsupervised learning and few-shot learning methods require normal samples during model training and operate within a texture close-set framework. They are limited to detecting texture types encountered in the training dataset. (d) The proposed zero-shot learning method stands out by not requiring any training cost. It operates under an open-set detection paradigm, enabling it to identify texture defects in open-world industrial scenarios.}
\label{fig2}
\end{figure*}

To address the aforementioned limitations and meet the practical requirements of modern inspection systems, the zero-shot learning-based texture anomaly detection paradigm is proposed as shown in Fig. 1(d). Unlike conventional methods mentioned above, the zero-shot learning approach eliminates the need for training data and can detect various textured defects in the open-set environment.
Recent zero-shot anomaly detection methods \cite{r34}\cite{r35} introduce large-scale pre-trained vision language models (VLMs) that have presented robust generalization capacities for detecting anomalies in novel textured surfaces.
Although these methods showcase some zero-shot recognition capabilities, they still depend on additional prior knowledge, such as base defect class information \cite{r55}, auxiliary training data \cite{r35}, or textual prompting \cite{r34}. Selecting appropriate prior knowledge demands significant human effort. Moreover, when prior knowledge is insufficient, such as in cases of lacking auxiliary data or inadequate textual prompts, their performance may degrade.

In contrast to conventional zero-shot learning techniques that depend on pre-existing prior knowledge, inspired by human visual perception \cite{r57}, we incorporate insights from both local and global perspectives for zero-shot texture anomaly detection \cite{r57}. In particular, two intrinsic prior patterns from the test texture image itself are directly extracted for anomaly detection, eliminating the need for any additional knowledge. The first is the local neighborhood prior, representing the most coherent region with similarity. The second is the global normality prior, employed to characterize the most typical normal pattern observed across the entire texture image. 

By effectively harnessing these two cues, we introduce Global-Regularized Neighborhood Regression (GRNR) to detect abnormal regions in texture images in zero-shot scenarios. For each local query within the texture image, we utilize a linear transformation of the local neighborhood to establish a regression model. This transformation is designed to minimize regression errors while adhering to global normality constraints. In regions characterized by normal textures, precise regression is achieved through neighbor priors exhibiting coherent similarity, which aligns with the requirements of global normality. However, abnormal defect regions deviate from neighborhood consistency and global normality constraints, leading to notable regression errors.
Additionally, we derive an approximate closed-form solution for regression to enable rapid inference. In summary, the proposed GRNR can effectively identify anomalies in novel textured surfaces without requiring additional data or prompts, demonstrating promising efficiency. The primary contributions of this paper are outlined as follows:

\begin{enumerate}
     \item This study introduces an efficient method GRNR for detecting texture anomalies in an open-set zero-shot learning manner without incurring any training costs. 

    \item  Our proposed GRNR involves extracting local neighborhood priors and global normal priors from texture images for query self-reconstructive regression. Additionally, we formulate an approximate closed solution for the optimal transformation of the regression for efficient inference.

    \item  A comprehensive benchmark dataset Texture Spectrum representing diverse industrial scenarios has been constructed in this study. Experimental results on the Texture Spectrum indicate that GRNR can accurately detect anomaly defects without requiring any training cost in zero-shot scenarios, even surpassing vanilla methods that rely on extensive sample training.
\end{enumerate}

The subsequent sections of this paper are organized as follows: Section II offers an overview of recent research on texture defect detection. Section III elaborates on our proposed methodology. Section IV provides a detailed description of our comprehensive experiments, and the final section presents conclusions and future perspectives.

\section{Related works}

Over the past two decades, various methods grounded in deep learning for texture anomaly detection have been proposed, and these methods have continued to evolve as the nature of supervision and data efficiency changes. 

\subsection{Supervised texture anomaly detection}

The earliest texture anomaly detection methods relied on a fully supervised paradigm, which necessitates large-scale labeled data for model training and covers a range of fine-grained tasks. For instance, in \cite{r1,r19}, anomaly classification technology was applied to solar panels and flat steel surfaces. To achieve more precise detection outcomes, object detection in visual tasks has also been utilized for anomaly bounding-box detection \cite{r4} \cite{r5}. For example, a multi-level feature fusion network for end-to-end strip anomaly defect detection was proposed in \cite{r3}.  The accurate pixel-level semantics segmentation task is common and has garnered significant attention. PGA-Net \cite{r7} addresses anomaly segmentation on various textured surfaces like strip steel surface \cite{r12}. However, these supervised methods require substantial manual effort in data collection and annotation, limiting their applicability to defect-closed environments, specifically to detect defects present during training. Consequently, they may not generalize well to various unknown types of anomalies that could potentially arise in open-world industrial settings.


\subsection{Unsupervised texture anomaly detection}

In response to the labor-intensive challenges of data labeling and the limitations of defect close-set detection, subsequent researchers have developed unsupervised methods for detecting anomalies in texture images. This method exclusively relies on unlabeled normal data to establish a normal distribution during training, which is then utilized during testing to identify unseen anomalous defects by calculating distribution discrepancy. The mainstream unsupervised techniques for detecting texture anomalies consist of reconstruction-based \cite{r20}, regression-based \cite{r21}, and density-based methods \cite{r22}. Reconstruction-based approaches typically employ models like autoencoders \cite{r52} for texture reconstruction. The reconstruction errors are employed to score anomalies. This methodology has garnered widespread popularity, like MSCDAE \cite{r11} and MS-FCAE \cite{r20}, both serving the purpose of texture anomaly segmentation. To enhance the autoencoder's reconstruction capabilities, the generative adversarial network \cite{r24} was further introduced, such as the AFEAN \cite{r25}, and the FMR-Net \cite{r26}. The regression-based method employs knowledge distillation with a teacher-student network for anomaly detection. In this approach, the student network learns only normal knowledge, and the discrepancies in predictions between the teacher and student networks are utilized to identify abnormalities. Various methods within this framework have been devised and explored for texture anomaly detection. These include multi-resolution distillation (KDAD)\cite{r28}, informative distillation (IKD)\cite{r21}, pre-trained feature mapping (PFM)\cite{r29}, and reverse distillation (RD4AD)\cite{r30}. Density-based methods directly characterize the distribution of normal samples. For instance, Padim \cite{r22} utilizes position-dependent multivariate Gaussian modeling to represent normal features, while GCPF \cite{r31} employs a mixture of Gaussian clustering distribution. Patchcore\cite{r32} establishes a normal feature library directly, determining abnormal defects through nearest neighbor search.

While these unsupervised approaches alleviate the burden of data labeling, they still require a substantial amount of normal samples for training. Furthermore, such methods are inherently texture-closed in that they are restricted to the textures encountered during training and cannot generalize to unseen texture types.

\subsection{Few$/$Zero-shot learning anomaly detection}

To devise a more labor- and data-efficient texture defect detection method, researchers have delved into few-shot and zero-shot learning methodologies. For few-shot anomaly detection, RegAD \cite{r18} employs an alignment-based meta-learning architecture to accomplish anomaly localization with a small number of samples. On the other hand,  FastRecon \cite{r17} leverages the few shot support samples for fast feature reconstruction. More recently, large-scale VLMs have exhibited robust zero-shot recognition capabilities across diverse visual tasks, facilitated by language modalities. Zero-shot anomaly detection methods have also started to utilize these VLMs. For instance, WinCLIP \cite{r34} utilizes the pre-trained VLM model CLIP alongside a window-based strategy to achieve anomaly segmentation based on text prompts. APRIL-GAN\cite{r35} introduces additional linear layers to align encoded image features with text features. Nevertheless, these existing zero-shot texture anomaly detection methods require carefully designed handcraft prompts \cite{r34} or auxiliary training datasets\cite{r35}, yet not generalized enough and may degrade with unsuitable prior information.

To tackle the aforementioned challenges, we introduce GRNR, a novel and efficient zero-shot texture anomaly defect detection scheme that requires no training. It operates without the need for additional prior information from training data or language prompts, instead extracting intrinsic prior directly from the texture image for fast and accurate abnormal defect detection.

\begin{figure*}[t]
\centerline{\includegraphics[width=150mm]{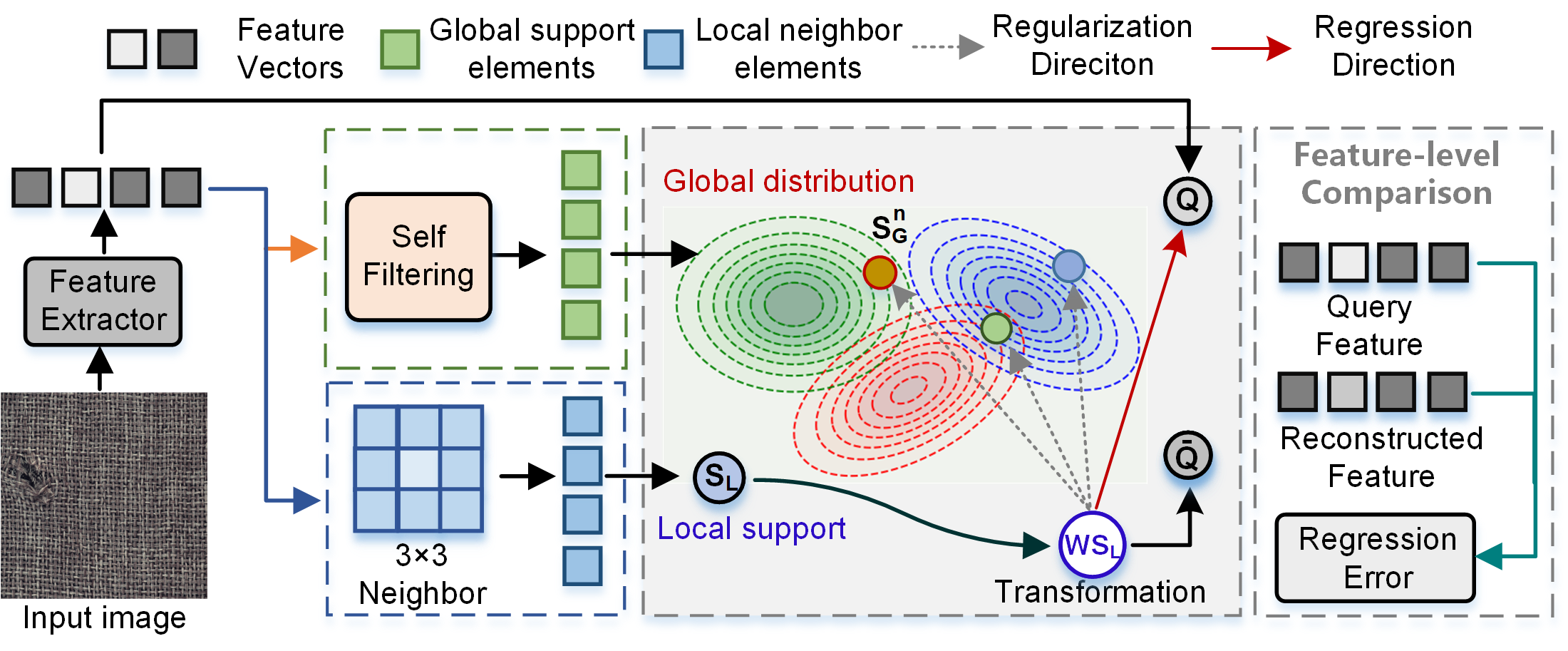}}
\caption[width=150mm]{
Method pipeline. The image is converted into patch-level embeddings using a feature extractor. Subsequently, global and local neighbor support elements $\left \{ \bm{S}_L, \bm{S}_G \right \}$, are derived through self-filtering and neighborhood operations. Following this, the global-regularized neighborhood regression process is executed to reconstruct individual query elements $\bm{Q}$. The optimal transformation $\tilde{\bm{W}}$ considers that the obtained $\bar{\bm{Q}}$ through $\tilde{\bm{W}}\bm{S}_L$ can achieve similarity with $\bm{Q}$(regression direction) while simultaneously adhering to the global normal distribution defined by 
$\bm{S}_G$(regularization direction). Anomaly scores are obtained through a feature-level comparison. 

}
\label{fig2}
\end{figure*}

\section{Methodology}

\subsection{Problem Definition}
The task of texture image anomaly detection entails the identification of anomalies within images. In this context, given an image $X$, the presence of specific anomalous defects is annotated using a pixel mask $M$, wherein 0 denotes normal pixels and 1 signifies abnormal pixels. The detection process can be categorized into two levels: image-level inspection and pixel-level segmentation. In pixel-level segmentation, the anomaly detection model is required to produce the corresponding anomaly score map $\mathcal{A}$. On the other hand, image-level inspection aims to describe the presence or absence of anomalies within the entire image based on $\mathcal{A}$.

As shown in Fig. 2, the proposed GRNR can generate the anomaly score map $\mathcal{A}$ directly from image $X$ without relying on any extra prior knowledge. In the following sections, the design details of GRNR will be subsequently elaborated.

\subsection{Feature embedding and support sampling}

For an image $X$, we adhere to the fashion of existing methods and employ a pre-trained convolutional neural network (CNN) $\mathcal{E}$ to embed the image into patch-level embedded feature representation $\phi_i =\mathcal{E} (X)$, 
where $i=\left \{1,..., \mathcal{H}  \right \} $, and $\mathcal{H}$ represents the number of feature hierarchies. 

For each hierarchy of the feature map in $\phi$, it is denoted as $\bm{F}\in \mathbb{R}^{C\times H\times W}$, where $C$, $H$, and $W$ represent the number of channels, height, and width of the feature map, respectively. For each element $\bm{Q}\in R^{1\times C}$ with the position of $\left ( h,w \right )$ in the obtained feature map $\bm{F}$, we can obtain its two corresponding support prior elements $\mathbf{S } = \left \{ \bm{S}_L, \bm{S}_G \right \}$ from both local and global perspectives, where $\bm{S}_L$ is the local neighbor support elements, consisting of the adjacent elements within a $m$-size local neighborhood $\mathcal{N}_Q^{\left ( h,w \right ) }$:
\begin{equation}
\begin{aligned}
\mathcal{N}_Q^{\left ( h,w \right ) }= \big \{ \left ( a,b \right )\mid \quad & a\in \left [ h-m,h+m \right ],\\  & b \in \left [ w-m,w+m \right ]  \big \}
\end{aligned}
\end{equation}

These elements represent the most similar coherent texture pattern of $\bm{Q}$, and $\bm{S}_L\in  R^{(4m^2+4m)\times C}$ of $\bm{Q}$ at position $\left ( h,w \right )$  can be obtained as: 
\begin{equation}
\bm{S}_L=\left \{ \bm{F}(a,b)\mid (a,b)\in \mathcal{N}_Q^{\left ( h,w \right ) },(a,b)\ne(h,w)\right \} 
\end{equation} 

On the other hand, $\bm{S}_G$ represents the most typical texture pattern in the entire textural image from a global perspective. To this end, we propose a self-filtering normality sampling method to obtain the $K$ most typical elements $\bm{S}_G \in R^{K\times C}$: 
\begin{equation}
\small
\bm{S}_G=\left \{ \mathrm{topK}\Big ( \underset{x,y}{\mathrm{argmin}}  \sum_{i=1}^{H} \sum_{j=1}^{W}\left \|\bm{F}(x,y)-\bm{F}(i,j)  \right \|^2   \Big ) \right \}
\end{equation}

In Eq. (3), the elements with the smallest sum of distances from the rest of the elements represent the most typical texture pattern in $\bm{F}$. This is due to the randomness and unpredictability of abnormality. Abnormal areas differ from normal texture and even from the abnormal areas themselves, while normal texture exhibits similar patterns, resulting in a smaller sum of distances.

\subsection{Global-Regularized Neighborhood Regression}

\subsubsection{Preliminary}
As mentioned in \cite{r36}\cite{r58}, the ridge regression is motivated by solving a linear least squares problem: using a linear transformation $\bm{W}$ of $\bm{X}$ to approximate $\bm{Y}$ that meets $\bm{WX}\approx\bm{Y}$:
\begin{equation}
\tilde{\bm{W}} =\underset{\bm{W}}{\mathrm{argmin} } \left \| \bm{Y}-\bm{W}\bm{X} \right \|^2+ \lambda (\bm{W})
\end{equation}
where the $\lambda (\cdot)$ is the penalty function of $\bm{W}$, typically utilizing the $L1$ or $L2$ regularization. If the penalty regularization term is ignored, the closed-form solution of Eq. (4) is as follows:
\begin{equation}
\tilde{\bm{W}} =\bm{Y}\bm{X}^{\mathrm{T } }\left ( \bm{X}\bm{X}^{\mathrm{T } } \right ) ^{-1}
\end{equation}
\subsubsection{Neighborhood regression}
Inspired by the above conclusions, we proposed the neighborhood regression that aims to utilize the linear transformation $\bm{W}$ of $\bm{S}_L$ to regress and reconstruct $\bm{Q}$. Our motivation stems from the observed fact about texture surfaces: due to the randomness and uncertainty of abnormal defects, abnormalities are difficult to be approximated by their surrounding elements, while regular normal textures can be reconstructed by regression through their surrounding similar elements.

\subsubsection{Global distributional regularization}
As pointed out in the reconstruction-based anomaly method\cite{r25, r26}, for the reconstruction of unknown anomaly defects, the desired outcome is that it can exhibit a pattern similar to normal texture. However, the phenomenon of over-constraint in the linear system mentioned above may cause $\bm{W}\bm{S}_L$ to be too close to $\bm{Q}$, leading it to deviate from the pattern distribution of normal texture.

To solve the above problem, a global distributional regularization is proposed as the penalty function in this paper. Specifically, by utilizing the acquired $\bm{S}_G$ representing the typical normal patterns, we model the distribution $f$ of normal texture as a Gaussian mixture density distribution centered on them:
\begin{equation}
f(\bm{W}\bm{S}_L\mid \bm{S}_G)= \sum_{n=1}^{N_c} \omega_n\mathcal{G} (\bm{W}\bm{S}_L \mid \bm{\mu}_{n} ,\bm{\Sigma}_{n } )
\end{equation}
where the $\mathcal{G} (\cdot\mid \bm{\mu}_{n} ,\bm{\Sigma}_{n } )$ denotes the $n$-th Gaussian distribution with the mean $\bm{\mu}_{n}$ and covariance $\bm{\Sigma}_{n}$ estimated by the $n$-th cluster center in $\bm{S}_G$, $\omega_n$ is the weight coefficient, and $N_c$ is the number of clusters. Therefore, based on the probability of the regression term $\bm{W}\bm{S}_L$ in the normal texture distribution, we propose to use the negative log probability as the global distribution regularization penalty function of the regression, forcing the $\bm{W}\bm{S}_L$ to be in the high probability area. Formally, this linear transformation $\bm{W}$ can be obtained as:
\begin{equation}
\tilde{\bm{W}} =\underset{\bm{W}}{\mathrm{argmin} }\left \{   \left \| \bm{Q}-\bm{W}\bm{S}_L \right \|^2- \beta \mathrm{log } \left ( f(\bm{W}\bm{S}_L\mid \bm{S}_G) \right ) \right \} 
\end{equation}
where the $\beta $ represents the weight of the penalty item.

\subsubsection{Closed-form solution fast approximation}

 The existence of the probability density function hinders the efficient solution of $\bm{W}$. Therefore, in this paper, we propose a fast approximation to the closed-form solution.  First, according to Jensen’s inequality, the upper bound of the second term of Eq. (7) can be written as: 
 \begin{equation}
 - \beta \mathrm{log } \left ( f(\bm{W}\bm{S}_L\mid \bm{S}_G) \right ) \le -\beta \sum_{n=1}^{N_c} \omega_n \mathrm{log }(\mathcal{G} (\bm{W}\bm{S}_L \mid \bm{\mu}_{n} ,\bm{\Sigma}_{n } ))
  \end{equation}
 
 Subsequently, for the $n$-th Gaussian distribution, its probability density function is given as:
 \begin{equation}
 \mathcal{G} (\bm{x}\mid \bm{\mu}_{n} ,\bm{\Sigma}_{n } )=\frac{1}{\sqrt{(2\pi)^C \left | \bm{\Sigma}_{n } \right | } }e^{-\frac{1}{2}(\bm{x}-\bm{\mu}_{n})^{\mathrm{T} }\bm{\Sigma}_{n }^{-1} (\bm{x}-\bm{\mu}_{n})} 
  \end{equation}
and its negative logarithmic form can be formulated as:
\begin{equation}
\begin{aligned}
  -\mathrm{log}( \mathcal{G} (\bm{x}\mid \bm{\mu}_{n} ,\bm{\Sigma}_{n } )) &= 
  \mathrm{log}( \sqrt{(2\pi)^C \left | \bm{\Sigma}_{n } \right | } )\\&+\frac{1}{2}(\bm{x}-\bm{\mu}_{n})^{\mathrm{T} }\bm{\Sigma}_{n }^{-1} (\bm{x}-\bm{\mu}_{n})
 \end{aligned}
 \end{equation}
 
Considering computational simplicity, we assume that each feature dimension is independent of the others. Consequently, we omit the calculation of covariances between features, resulting in a symmetric and positive definite diagonal covariance matrix 
 $\bm{\Sigma}_{n }$. This leads us to the following transformation:
 \begin{equation}
 \begin{aligned}
(\bm{x}-\bm{\mu}_{n})^{\mathrm{T} }\bm{\Sigma}_{n }^{-1} (\bm{x}-\bm{\mu}_{n})&=(\bm{x}-\bm{\mu}_{n})^{\mathrm{T} }\bm{B}_{n }^{\mathrm{T}}\bm{B}_{n }  (\bm{x}-\bm{\mu}_{n})\\
&=
\left [ \bm{B}_{n }(\bm{x}-\bm{\mu}_{n})\right ] ^{\mathrm{T} }
\left [ \bm{B}_{n }(\bm{x}-\bm{\mu}_{n})\right ]\\
&=
\left \| \bm{B}_{n }(\bm{x}-\bm{\mu}_{n}) \right \|^2 
 \end{aligned}
 \end{equation}
where $\bm{B}_{n }$ is an invertible matrix that meets $\bm{B}_{n }^{\mathrm{T}}\bm{B}_{n }=\bm{\Sigma}_{n }^{-1} $. According to the principle of compatibility, we can represent the upper bound of Eq. (11) as follows:
 \begin{equation}
  \begin{aligned}
\left \| \bm{B}_{n }(\bm{x}-\bm{\mu}_{n}) \right \|^2 &\le \left \|\bm{B}_{n }\right \|_F\left \|\bm{x}-\bm{\mu}_{n} \right \|^2 \\
&=\sqrt{\mathrm{Tr}(\bm{B}_{n }^{\mathrm{T}}\bm{B}_{n }) } \left \|\bm{x}-\bm{\mu}_{n} \right \|^2\\
&=\sqrt{\mathrm{Tr}(\bm{\Sigma}_{n }^{-1}) } \left \|\bm{x}-\bm{\mu}_{n} \right \|^2
 \end{aligned}
 \end{equation}
where the $\left \| \cdot \right \|_F$ is the Frobenius norm, and $\mathrm{Tr}(\cdot)$ represents the trace of matrix.
Combining Eq. (8), (10), and (12), we derive the following expression:
 \begin{equation}
   \begin{aligned}
 &-\beta \sum_{n=1}^{N_c} \omega_n \mathrm{log }(\mathcal{G} (\bm{W}\bm{S}_L \mid \bm{\mu}_{n} ,\bm{\Sigma}_{n } )) \\
 &\le\beta \sum_{n=1}^{N_c} \omega_n\big (\mathrm{log}( \sqrt{(2\pi)^C \left | \bm{\Sigma}_{n } \right | }+\sqrt{\mathrm{Tr}(\bm{\Sigma}_{n }^{-1}) } \left \|\bm{W}\bm{S}_L -\bm{\mu}_{n} \right \|^2 \big ) \\
 &=\sum_{n=1}^{N_c} (\mathbf{C}_1(\bm{\Sigma}_{n } )+\mathbf{C}_2(\bm{\Sigma}_{n}^{-1})  \left \|\bm{W}\bm{S}_L -\bm{\mu}_{n} \right \|^2 )
  \end{aligned}
  \end{equation}
where the $\mathbf{C}_1(\bm{\Sigma}_{n } ))$,$\mathbf{C}_2(\bm{\Sigma}_{n}^{-1})$ denote the constants related to the covariance matrix $\bm{\Sigma}_{n }$ and $\omega_n$. 
Substituting Eq. (13) into Eq. (7) yields the upper bound of the optimization objective as follows:
 \begin{equation}
    \begin{aligned}
\tilde{\bm{W}} =\underset{\bm{W}}{\mathrm{argmin} }\Big \{  & \left \| \bm{Q}-\bm{W}\bm{S}_L \right \|^2 \\
&+\mathbf{C}_1(\bm{\Sigma}_{n } ))+\sum_{n=1}^{N_c}\mathbf{C}_2(\bm{\Sigma}_{n}^{-1})  \left \|\bm{W}\bm{S}_L -\bm{\mu}_{n} \right \|^2 \Big \} 
  \end{aligned}
  \end{equation}
  
Note that $\mathbf{C}_1$ is independent of $\bm{W}\bm{S}_L$, which can be disregarded, while $\mathbf{C}_2$ essentially controls the regularization intensity of different Gaussian clusters, we can consistently represent it as $\eta $ for computational convenience. More specifically, $\bm{\mu}_{n}$ denotes the $n$-th Gaussian clustering center within $\bm{S}_G$. Given that $\bm{S}_G$ already embodies representative patterns, we can directly employ the $\bm{S}_G^n$ to signify $\bm{\mu}_{n}$, in which case $N_c=K$. Finally, the resulting optimization objective is as follows: 
 \begin{equation}
\tilde{\bm{W}} =
\underset{\bm{W}}{\mathrm{argmin} }  \left \{ \| \bm{Q}-\bm{W}\bm{S}_L  \|^2+  \eta\sum_{n=1}^{K}  \left \|\bm{W}\bm{S}_L-\bm{S}_G^{n} \right \| ^2 \right \} 
\end{equation}

Subsequently, let $\bm{Q}^\ast$ and $\bm{S}^\ast$ represent the column-wise enhanced form as follows: 
\begin{equation}
\left\{\begin{matrix}
\bm{Q}^\ast =[\bm{Q},\sqrt{\eta} \bm{S}_G^{1},...,\sqrt{\eta} \bm{S}_G^{K}]  \\
\bm{S}^\ast =[\bm{S}_L,\overbrace{\sqrt{\eta} \bm{S}_L,...,\sqrt{\eta} \bm{S}_L}^{K}]
\end{matrix}\right.
\end{equation}
and Eq. (15) can be written as:
\begin{equation}
\tilde{\bm{W}} =\underset{\bm{W}}{\mathrm{argmin} } \left \| \bm{Q}^\ast-\bm{W}\bm{S}^\ast \right \|^2
\end{equation}

According to the conclusion derived from Eq.(5), the closed solution of the optimal transformation $\tilde{\bm{W}}$ is as follows:
\begin{equation}
\begin{aligned}
\tilde{\bm{W}} &=\bm{Q}^\ast \bm{S}^{\ast\mathrm{T } }\left ( \bm{S}^\ast\bm{S}^{\ast\mathrm{T } } \right ) ^{-1}\\
&=\bigg ( \bm{Q}\bm{S}_L^{\mathrm{T }}+\eta \sum_{n=1}^{K}\bm{S}_G^{n}\bm{S}_L^{\mathrm{T }}  \bigg ) \bigg (\bm{S}_L\bm{S}_L^{\mathrm{T }}+K\eta\bm{S}_L\bm{S}_L^{\mathrm{T }}  \bigg )^{-1}
\end{aligned}
\end{equation}

\subsection{Anomaly scoring with regression error}

Upon acquiring the optimal transformation, the regression model can be utilized to generate the reconstructed query, which is then compared with the original query to derive its anomaly score. For query $\bm{Q}(h,w)$ at the position $(h,w)$ of $i$-th feature-hierarchy, its anomaly score is defined as follows:
\begin{equation}
\mathcal{A}^i(h,w)= \left \| \bm{Q}(h,w) -\bar{\bm{Q}}(h,w) \right \| ^2
\end{equation}
where $\bar{\bm{Q}}$ is the reconstructed query obtained using the optimal transformation $\tilde{\bm{W}}\bm{S}_L$. Subsequently, the anomaly  maps at different feature hierarchies are integrated to derive the final anomaly score map $\mathcal{A}$:
\begin{equation}
\mathcal{A}=\prod_{i}\Psi (\mathcal{A}^i)
\end{equation}
where  $\Psi$ denotes the upsampling interpolation operation, designed to standardize the anomaly score maps from various feature levels to a consistent spatial dimension. The maximum value of $\mathcal{A}$ is taken as the image-level anomaly score.

\begin{figure*}[t]
\centerline{\includegraphics[width=180mm]{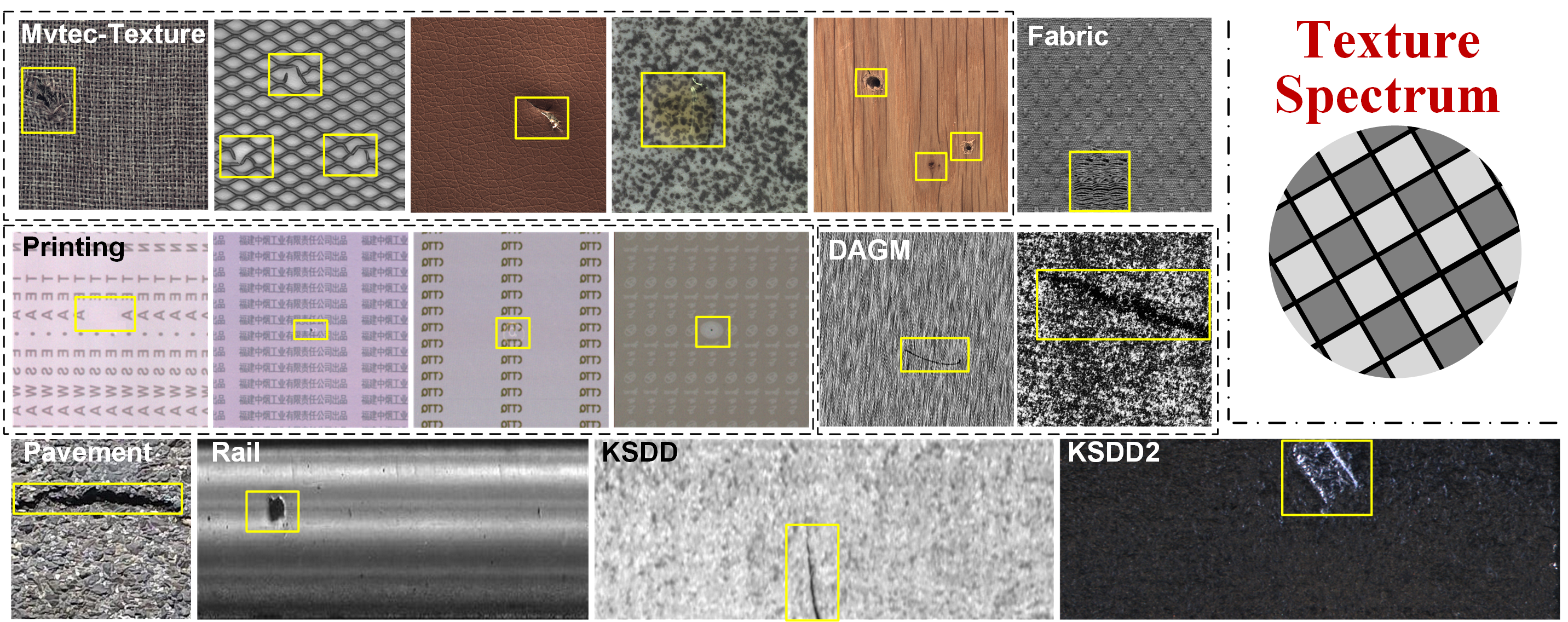}}
\caption[width=180mm]{
Some representative samples from the constructed Texture Spectrum dataset, which encompasses a diverse array of textured surfaces featuring both homogeneous and non-homogeneous textures.  Moreover, it includes various defect types such as cracks, holes, scratches, and others.
}
\label{fig2}
\end{figure*}

\section{Experiment}

This section offers a comprehensive experimental validation of the proposed GRNR methodology. To accomplish this objective, this study introduces the Texture Spectrum, an extensive benchmark encompassing diverse industry scenarios. Additionally, this section presents ablation studies on GRNR. Furthermore, the distinct advantages of GRNR over existing methods were also analyzed.

\begin{figure}[!t]
\centering
\includegraphics[width=88mm]{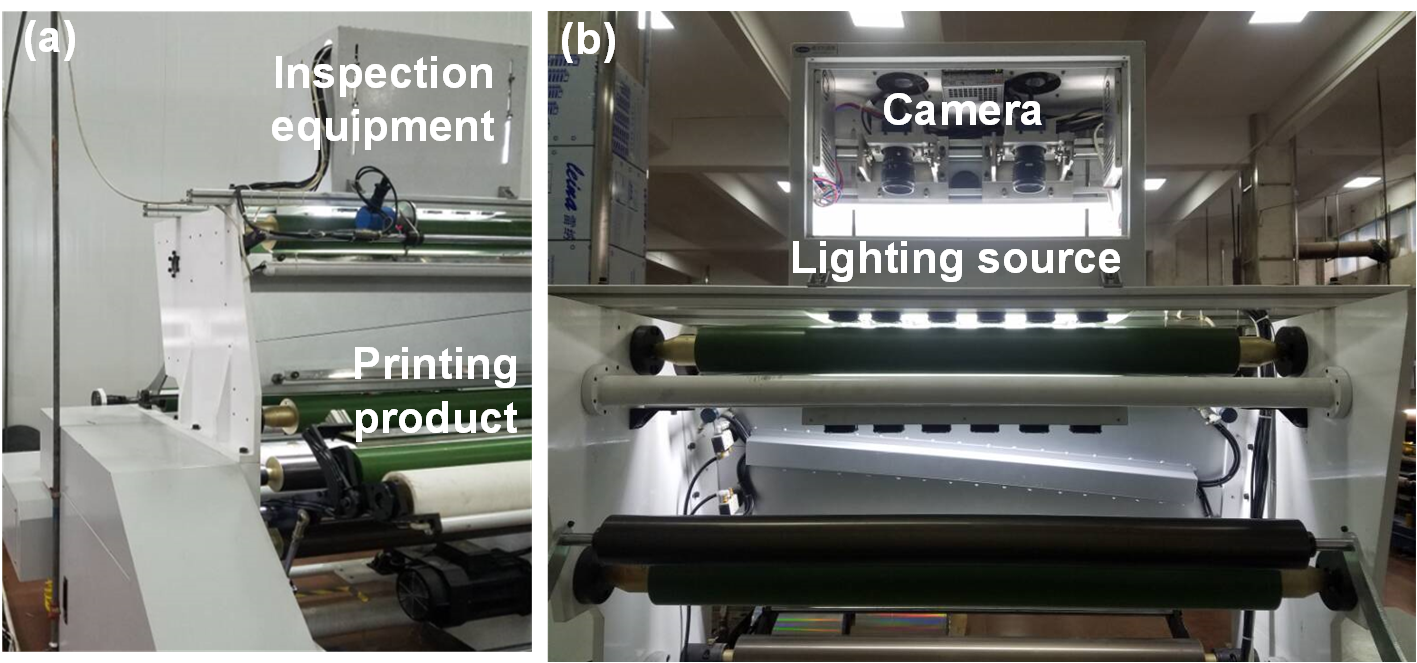}
\caption{(a). Printing product production site. (b). Our AOI instruments for printing defect inspection.}
\label{fig_1}
\end{figure}

\subsection{Established datasets}

As illustrated in Fig. 3, a novel and comprehensive texture benchmark named Texture Spectrum was proposed in this study for industrial defect detection. This dataset contains texture defect data collected from eight industrial environments, making it a comprehensive benchmark for fully evaluating defect detection methods. The detailed composition of the texture spectral dataset is outlined below:

(1)\textbf{Mvtec-AD texture} \cite{r8}. The Mvtec AD-Texture dataset is a widely utilized benchmark for validating unsupervised anomaly detection techniques. It encompasses five distinct real-world texture categories, including leather, wood, carpet, grid, and tile, and encompasses a range of defect types within these categories.

(2) \textbf{Pavement} \cite{r9}. The pavement dataset stands for the texture defect detection requirements specific to the construction industry. It encompasses texture data from cement, gravel, and asphalt pavements, with a primary emphasis on detecting cracks, which are the predominant defect types found in pavement surfaces.

(3) \textbf{Rail} \cite{r44}. This dataset comprises defects observed on the surface of an express railway. Due to variations in driving and lighting conditions, the dataset showcases diverse surface properties, thereby providing comprehensive validation of the model's performance.

(4)\textbf{Fabric} \cite{r45}. The Fabric data set representing the textile industry is derived from the Industrial Automation Research Laboratory of the Department of Electrical and Electronic Engineering of the University of Hong Kong. There are 3 fabric patterns available: star, dot, and box. Each pattern comprises 6 distinct defect types, which include broken end, hole, knot, netting multiple, thick bar, and thin bar.

(5) \textbf{DAGM} \cite{r47}. The DAGM dataset is a comprehensive artificially synthesized benchmark specifically designed for texture defect detection. It encompasses various types of defects, including scratches and spots. Following \cite{r48}, we have selected 7 representative textures from this dataset for validation.

\begin{figure}[!t]
\centering
\includegraphics[width=88mm]{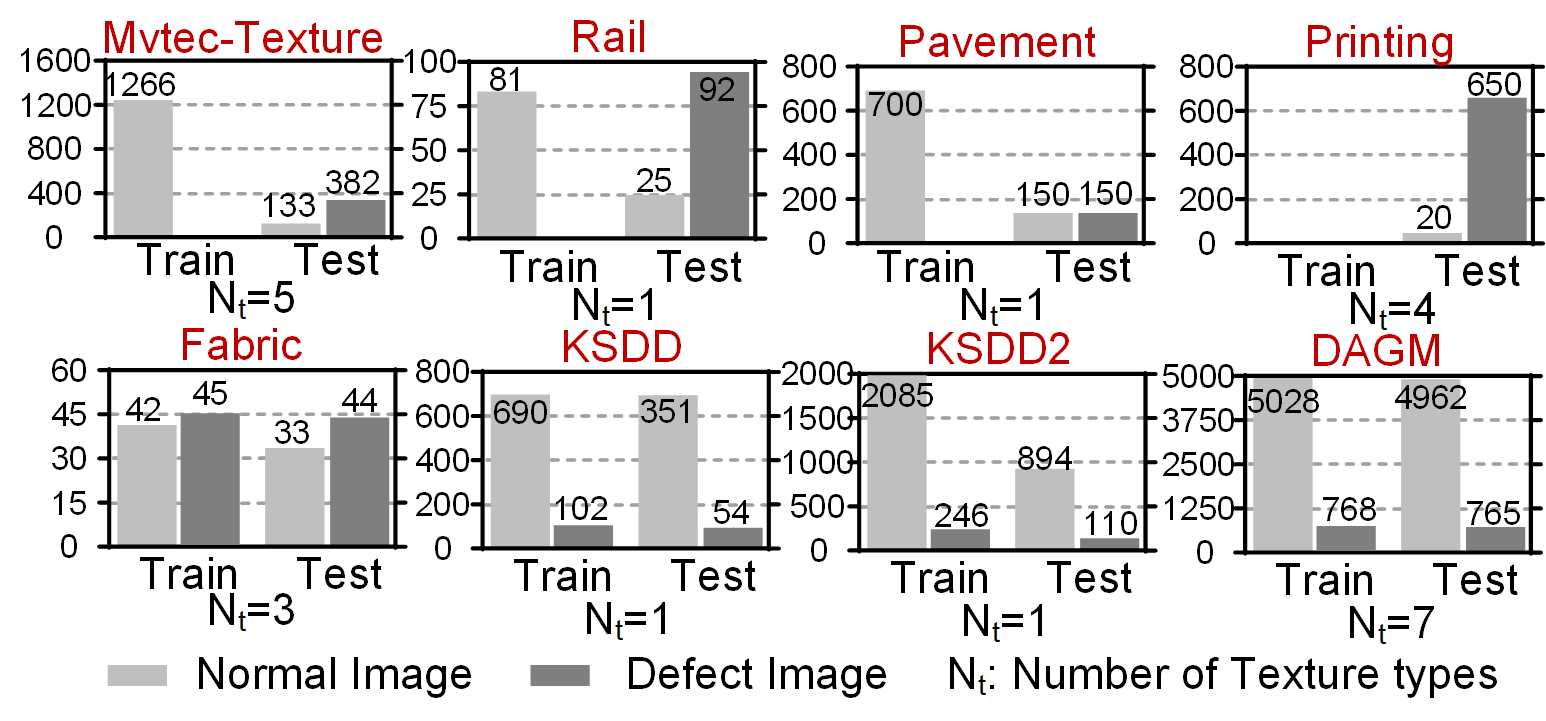}
\caption{The detailed configuration of each sub-dataset in Texture Spectrum, $N_t$ represents the number of texture types.}
\label{fig_1}
\end{figure}

(6) \textbf{KSDD}\cite{r6}. The KSDD dataset comprises images of faulty electrical commutators, predominantly exhibiting defects in the form of cracks.

(7) \textbf{KSDD2}\cite{r50}. The KSDD2 dataset is acquired through a visual inspection system operating on a real industrial production line. It encompasses a diverse range of anomalies, presenting a particular challenge due to defects that closely resemble normal appearances.

(8) \textbf{Printing}. This self-built dataset is sourced from real printed product surfaces, depicted in Fig. 4. Differing from the previously mentioned datasets, it has 4 distinct anisotropic patterns and contains complex textures such as text.

The training and testing set details for the eight sub-datasets, encompassing the count of normal and defective samples, along with texture categories, are illustrated in Fig. 5.

 \subsection{Baseline methods}
 

In this investigation, we present baseline methods featuring varying degrees of supervision to assess the performance of different approaches across various levels of human labor and data utilization. Specifically, for supervised methods that rely on a substantial number of labeled defect images for training, we selected advanced semantic segmentation methods PSPNet \cite{r38}, DeeplabV3+ \cite{r39}, and UperNet \cite{r40}. We compared these with unsupervised methods that require a large number of unlabeled normal images for training, including SPADE \cite{r41}, Padim \cite{r22}, Patchcore \cite{r32}, RD4AD \cite{r30}, and Draem \cite{r42}. Additionally, we evaluated few-shot learning methods such as RegAD \cite{r18}, Patchcore \cite{r32}, and FastRecon \cite{r17}, which necessitate only a small number of normal samples for training. Furthermore, we assessed the latest zero-shot learning methods WinClip \cite{r34}, SAA \cite{r43}, and April-GAN \cite{r35}, driven by pre-trained VLMs and text prompts. For all methods mentioned above, their corresponding publicly available codes are utilized for evaluations.

\subsection{Task settings}

In the experiments, we established diverse task settings to facilitate comparisons across methods featuring varying levels of supervision natures. We conducted a thorough comparison of supervised, unsupervised, few-shot, and zero-shot methods on the Fabric, KSDD, KSDD2, and DAGM datasets, which contain abnormal samples with corresponding labels in the training set. This situation falls under the defect close-set detection task setting. On the other hand, since the training sets of the MvtecAD-Texture, Rail, and Pavement datasets solely consist of normal training samples, supervised methods are not applicable. Hence, we applied unsupervised, few-shot, and zero-shot learning methods for comparison, which falls under the texture close-set detection task setting. Furthermore, in the case of the self-constructed Printing dataset, all samples were integrated into the test set, presenting the open-set detection scenario used to evaluate the effectiveness of the zero-shot learning approach. The specific task settings in the experiment are detailed in Table I.


\begin{table}
\caption{Experimental task settings}
\label{table}
\setlength{\tabcolsep}{3pt}
\centering
\begin{tabular}{p{88mm}}
${\includegraphics[width=88mm]{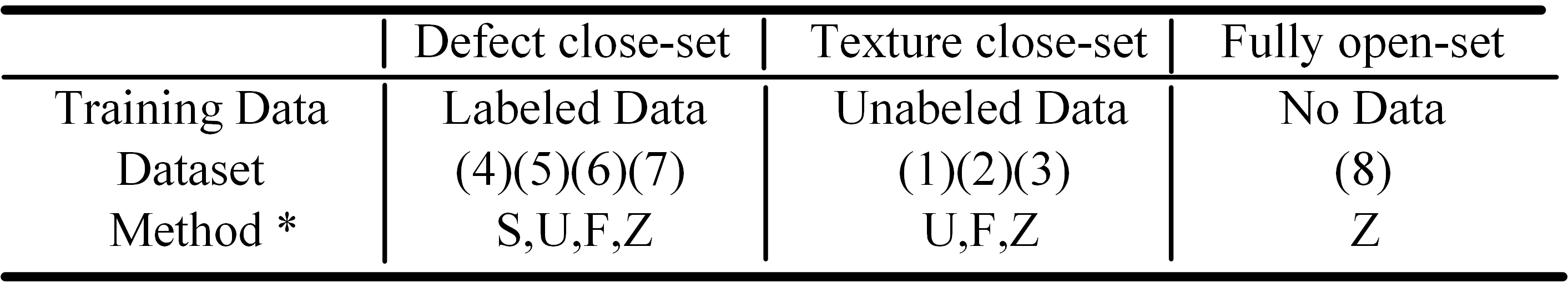}}$
\end{tabular}
\begin{tablenotes}
        \footnotesize
        \item[1]*: The S, U, F, and Z refer to supervised, unsupervised, few-shot, and zero-shot learning methods respectively.
\end{tablenotes}
\renewcommand{\thetable}{\Roman{I}}
\label{table_button}
\end{table}

\subsection{Evaluation Metrics}

When assessing the alignment between model predictions and actual labels, we employ True Positives (TP) and True Negatives (TN) to denote accurately predicted positive and negative instances. Conversely, False Positives (FP) and False Negatives (FN) represent misclassifications, where positive samples are erroneously labeled as negative, and negative samples are misclassified as positive. It's essential to note that positive and negative samples correspond to abnormal defective and normal instances, respectively.

\begin{table*}
\caption{Comparative results of pixel-level localization AUROC and F1-measure on the defect close-set task.}
\label{table}
\setlength{\tabcolsep}{3pt}
\centering
\begin{tabular}{p{180mm}}
${\includegraphics[width=180mm]{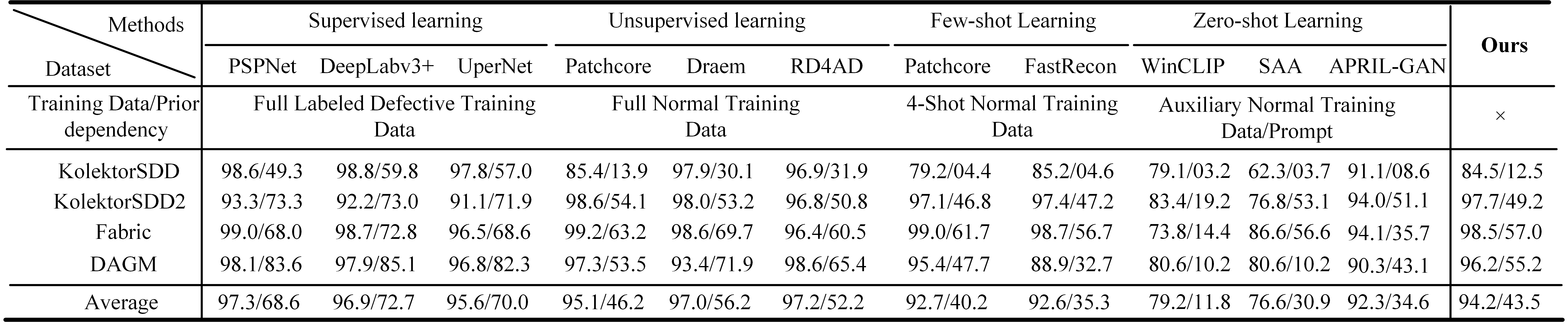}}$
\end{tabular}
\label{table_button}
\end{table*}

\begin{table*}
\caption{Comparative results of image/pixel-level detection and localization AUROC on the texture close-set task.}
\label{table}
\setlength{\tabcolsep}{3pt}
\centering
\begin{tabular}{p{180mm}}
${\includegraphics[width=180mm]{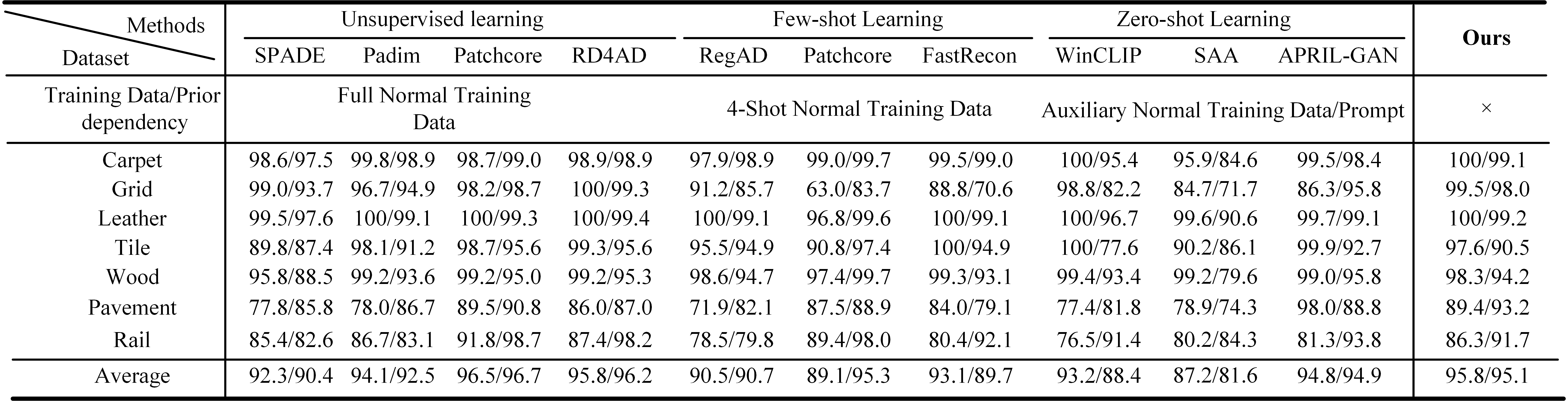}}$
\end{tabular}
\label{table_button}
\end{table*}

This study employs evaluation metrics in two scenarios: with a specified threshold and without any threshold. In binary segmentation scenarios with a specified threshold, we utilize the $F1-\mathrm{Measure}$ as the comprehensive evaluation metric:
\begin{equation}
F1-\mathrm{Measure} =\frac{2\times \mathrm{Recall} \times \mathrm{Precision}}{\mathrm{Precision} + \mathrm{Recall}}
\end{equation}
where the \emph{Precision} and \emph{Recall} are defined as follows:
\begin{equation}
\begin{aligned}
\mathrm{Precision} &=\frac{TP}{TP+FP} \times 100 \% \\
\mathrm{Recall} &=\frac{TP}{TP+FN} \times 100 \% 
\end{aligned}
\end{equation}
 Furthermore, under the threshold-free condition, the area under the curve
of receiver-operating-characteristic (AUROC) is also utilized to assess the performance of the compared methods:
\begin{equation}
AUROC=\int_{0}^{1} \mathrm{TPR}\cdot d(\mathrm{FPR})
\end{equation}
where the $\mathrm{TPR}$ and $\mathrm{FPR}$ are defined as follows:
\begin{equation}
\begin{aligned}
\mathrm{TPR} &=\frac{TP}{TP+FN} \times 100 \% \\
\mathrm{FPR} &=\frac{FP}{TN+FP} \times 100 \% 
\end{aligned}
\end{equation}
For both the $F1-\mathrm{Measure}$ and AUROC metrics, higher values indicate superior performance.

\subsection{Implementation Details}

During the experiments, each image underwent resizing to a resolution of 320 × 320 and normalization using the mean and standard deviation derived from the ImageNet dataset. To mitigate edge effects, we employed a 256×256 center cropping approach and applied Gaussian filtering with a sigma value of 4 to the anomaly score map. The Wide-Resnet50 is used as the feature extractor by default, and the hierarchical feature maps are adopted from $i=\left \{2, 3 \right \} $ level. The model's hyperparameters, neighborhood size $m$, number of the global support elements $K$, and regularization strength $\eta$, were set to default values of 1, 40, and 5, respectively. All experiments were carried out on a computer equipped with a 12GB NVIDIA GTX3060 GPU.

\subsection{Comparative experimental results}

\subsubsection{Results on defect close-set task}

For the defect close-set detection task, a systematic comparison was conducted between the proposed method and supervised, unsupervised, and few-shot methods. The detailed comparison results on the Fabric, DAGM, KSDD, and KSDD2 datasets are presented in Table II. The findings indicate that the supervised method exhibits the highest performance, particularly in terms of the $F1-\mathrm{Measure}$ metric. This advantage stems from the close-set  nature of defects in both the training and test sets, enabling the supervised method to accurately identify defects present in the training data. However, this advantage necessitates the collection of a substantial number of defect samples along with meticulous pixel-level annotations, which can be labor-intensive in industrial settings.

Unsupervised methods, while avoiding the need for annotations, still require significant manual effort to gather normal samples. Additionally, the detection performance of few-shot methods like Patchcore and FastRecon, which utilize a small number (4-shot) of normal samples, shows varying degrees of decline on the KSDD and DAGM datasets. Existing zero-shot learning methods that rely on text prompts, such as WinCLIP and SAA, exhibit significant performance degradation in terms of AUROC and F1, whereas APRIL-GAN's performance remains acceptable but with the cost of reliance on auxiliary training data set.

On the contrary, our proposed method achieves average $F1-\mathrm{Measure}$ and AUROC results of 43.5 
and 94.2, respectively. These metrics are comparable to the performance of unsupervised methods and surpass all existing few-shot learning methods and other zero-shot learning methods. Notably, our method achieves this level of performance without any prior prompts or incurring additional training costs, demonstrating its competitive performance.
\begin{figure*}[t]
\centerline{\includegraphics[width=160mm]{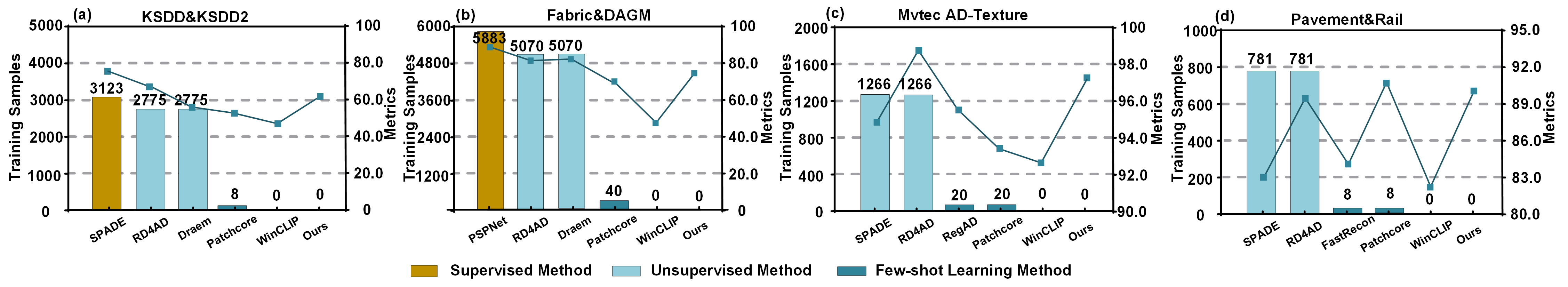}}
\caption[width=160mm]{
Histograms of training sample number comparison (left y-axis) and line chart of detection performance comparison (right y-axis) of different sub-datasets. It is worth noting that the performance of the model is represented by the average metrics reported in Tables II and III.
}
\label{fig2}
\end{figure*}

\begin{table}
\caption{Comparative results of pixel-level localization AUROC and F1-measure on the open-set task.}
\label{table}
\setlength{\tabcolsep}{3pt}
\centering
\begin{tabular}{p{88mm}}
${\includegraphics[width=88mm]{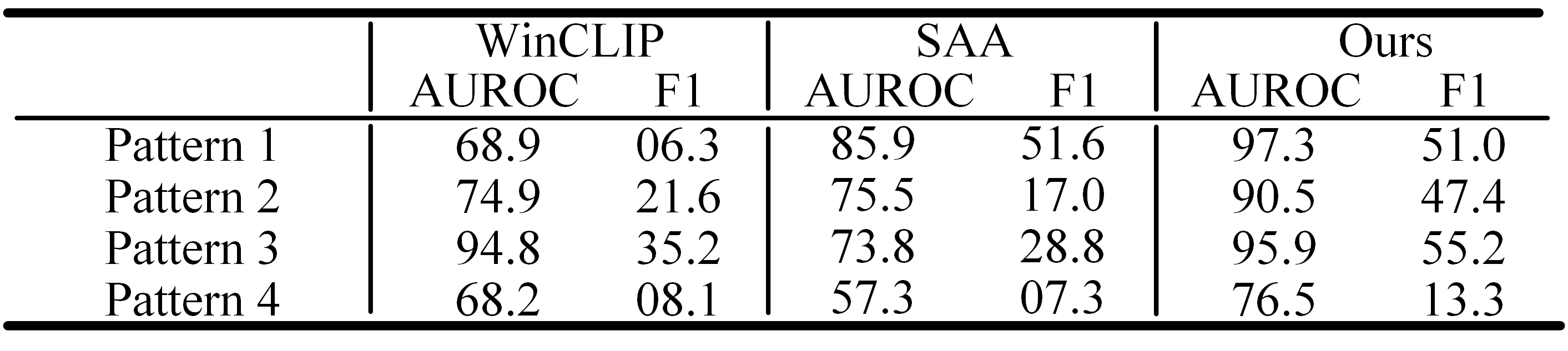}}$
\end{tabular}
\label{table_button}
\end{table}

\subsubsection{Results on texture close-set task}
Table III presents the comparison results for the texture closure detection task, encompassing unsupervised, few-shot, and zero-shot methods. The comparison reveals that existing unsupervised methods yield commendable results but necessitate a large volume of normal samples and extended training periods. With the reduction in training samples, the few-shot method exhibits varying degrees of performance decline. Similarly, other zero-shot learning methods also experience performance degradation in the absence of training samples.

Notably, our proposed zero-shot method achieves comparable outcomes to unsupervised methods trained with extensive data, surpassing SPADE and Padim, and only underperforms Patchcore and RD4AD. Compared to other zero-shot methods, our approach significantly outperforms WinCLIP and SAA, and marginally surpasses April-GAN. Nevertheless, it is important to note that April-GAN requires auxiliary training, while our method leverages inherent prior information directly from the test image itself, requiring no training, additional auxiliary data, or textual prompting. These experimental findings underscore the efficacy of our approach in achieving superior detection performance under highly efficient data and labor conditions.

\subsubsection{Results on open-set task}
For the open set detection task, we conducted a comparison between our proposed method and existing training-free zero-shot learning techniques WinCLIP and SAA using the self-constructed printing dataset to assess the detection capability without prior knowledge of the defect. The outcomes of these experiments are delineated in Table IV. The findings illustrate that our approach achieves superior performance than WinCLIP and SAA methods that rely on pre-trained VLM and text prompts. Particularly, when compared to SAA, our method demonstrates a notable enhancement of +16.9/+15.6 in average segmentation AUROC and $F1-\mathrm{Measure}$ over four textural patterns.
\begin{table}
\caption{ Comparison of computational complexity of different models}
\label{table}
\setlength{\tabcolsep}{3pt}
\centering
\begin{tabular}{p{88mm}}
${\includegraphics[width=88mm]{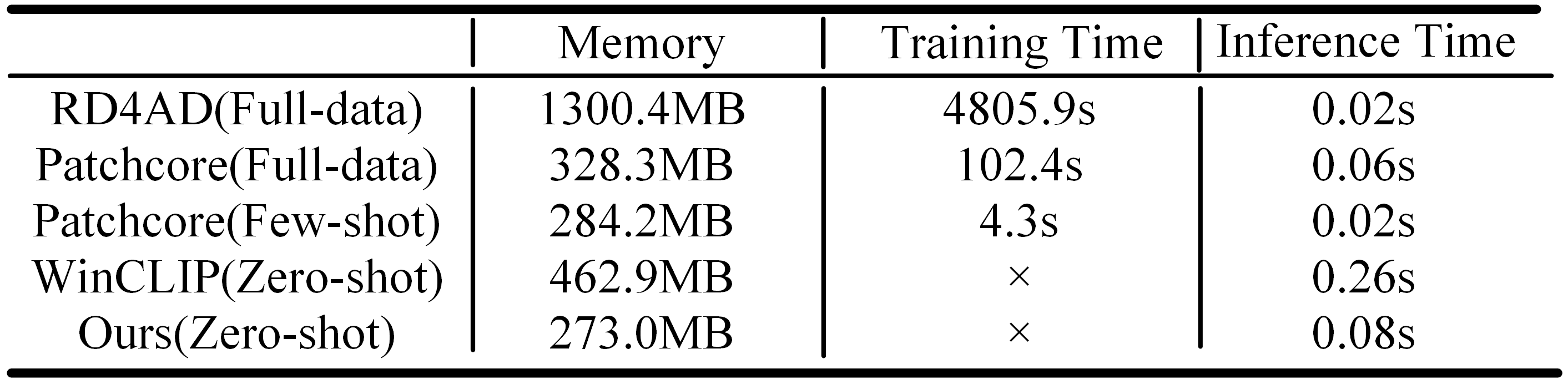}}$
\end{tabular}
\label{table_button}
\end{table}

\begin{figure*}[t]
\centerline{\includegraphics[width=160mm]{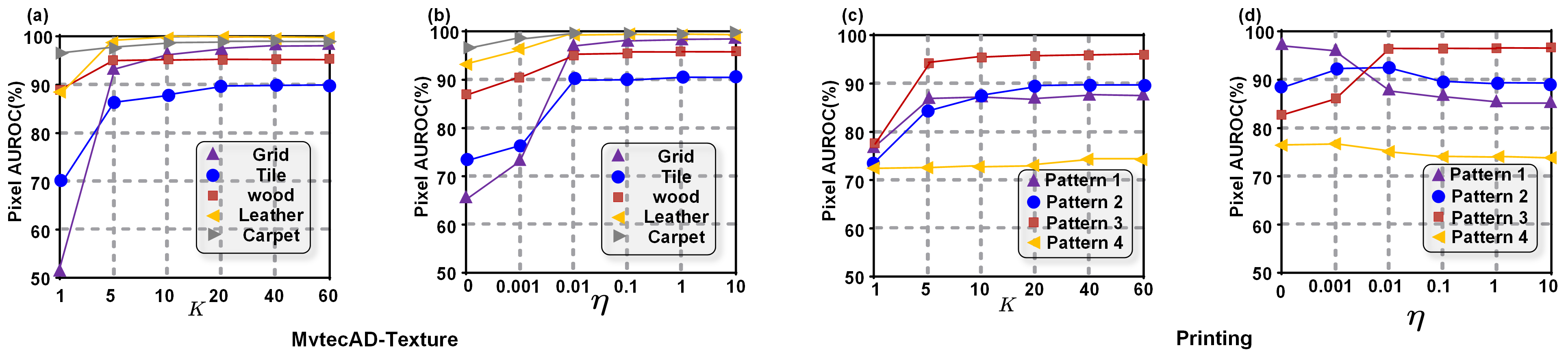}}
\caption[width=160mm]{
(a)-(b): The influence of the number of global support elements $K$ and the regularization strength $\eta$ on the model's performance on the MvtecAD-Texture dataset. (c)-(d): The influence of the number of global support elements $K$ and the regularization strength $\eta$ on the model's performance on the Printing  dataset.
}
\label{fig2}
\end{figure*}

\begin{figure*}[t]
\centerline{\includegraphics[width=170mm]{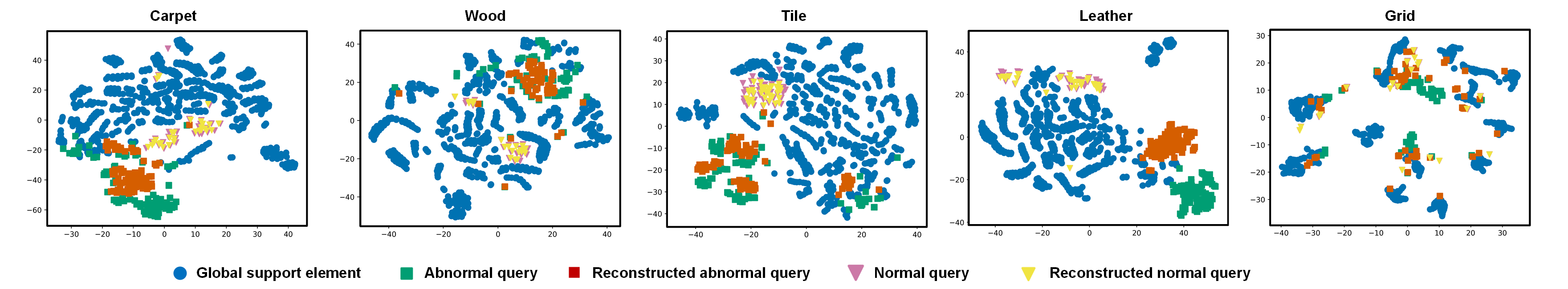}}
\caption[width=160mm]{
T-SNE\cite{r51} visualization results of the regression behavior of the model on the MvtecAD-Texture dataset.
}
\label{fig2}
\end{figure*}

\subsection{Cost analysis}

In this section, we delve into a cost analysis of the proposed method, focusing on data efficiency and computational efficiency.

Firstly, our method is versatile in tackling detection tasks related to defect close-set, texture close-set, and open-set, unlike other supervised methods that are limited to close-set detection. For example, supervised methods are suitable for detection tasks in defect close-set, while unsupervised and few-shot methods are exclusively applicable to texture close-set. This underscores the broad applicability of our approach across different detection scenarios.

Secondly, we delve into the data efficiency of the proposed method. As depicted in Fig. 6, we present an analysis of the model's performance and its dependency on data across different datasets. On the KSDD, KSDD2, Fabric, and DAGM datasets, supervised methods leveraging pixel-level annotations perform best, followed by unsupervised methods. However, as illustrated in Fig. 6(a)-(b), these approaches necessitate a substantial volume of training data. On the MvtecAD-Texture, Pavement, and Rail datasets, where large volumes of training data drive the unsupervised methods exhibit superior performance while few-shot methods show a decline as depicted in Fig. 6(c)-(d). In contrast, even without using any training data, our method demonstrates robust performance, which maintains detection ability comparable to unsupervised methods.  Considering the human labor and computational costs incurred by other methods utilizing large-scale datasets, our approach remains highly competitive.

Furthermore, we delve into the computational complexity of our proposed approach on the Carpet category in MvtecAD-Texture.  As delineated in Table V, in comparison to existing methods, our method excels by not necessitating any training time. In contrast, the unsupervised RD4AD and Patchcore methods require extensive model training time or substantial memory utilization. Moreover, compared with the zero-shot method WinCLIP, which requires the pre-trained VLM CLIP, our proposed method exhibits better inference efficiency and less memory usage, necessitating only 80 milliseconds for inference. Overall, the proposed method demonstrates high computational efficiency.

\begin{table}
\caption{The impact of local neighbor size on the segmentation\\ AUROC and inference time of the GRNR model.}
\label{table}
\setlength{\tabcolsep}{3pt}
\centering
\begin{tabular}{p{75mm}}
${\includegraphics[width=75mm]{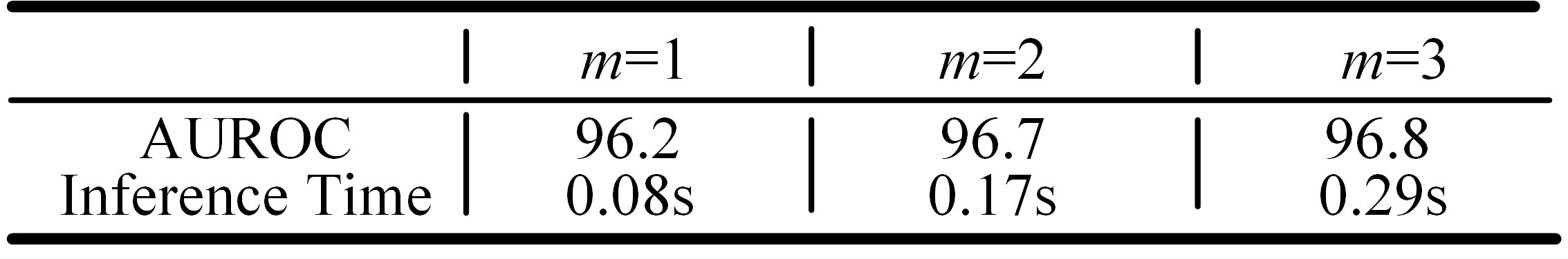}}$
\end{tabular}
\label{table_button}
\end{table}

\subsection{Ablation experiment}

In this section, we delve into the key hyper-parameters of the proposed method. These hyper-parameters encompass the size of the local neighbor $m$, the number $K$ of global support elements $\bm{S}_G$, and the regularization strength $\eta$.

\subsubsection{The effect of local neighbor size}
We evaluate the influence of the neighborhood size $m$ on MvtecAD-Texture, as detailed in Table VI. GRNR leverages the local neighborhood elements of the query for reconstruction and regression. A larger local neighborhood can encompass more local support elements exhibiting similar coherence, thereby enhancing the model's performance. However, concurrently, a larger neighborhood size leads to increased computational complexity in the regression model. Given the marginal performance enhancement and substantial rise in computational complexity, we opted for 
$m=1$ by default.

\subsubsection{The influence of global support number}

The influence of the number of global support elements $K$ on the model's performance across various categories is depicted in Fig. 7(a) and (c). As $K$ increases, the performance of GRNR gradually improves across all textures, with the most significant enhancement observed in the Grid category within the Mvtec AD-Texture dataset. This phenomenon is attributed to the fact that global support elements encapsulate the most typical patterns found in texture images. When the number of elements is small, they may not encompass all the normal patterns of the texture, leading to suboptimal performance. Increasing $K$ enriches the normal pattern information within $\bm{S}_G$ and provides more accurate distribution regularization, bringing enhanced detection performance. However, it is important to note that higher values of $K$ result in increased computational complexity. Hence, $K=40$ is chosen as the default option.

\subsubsection{The impact of regularization strength}

\begin{figure*}[t]
\centerline{\includegraphics[width=180mm]{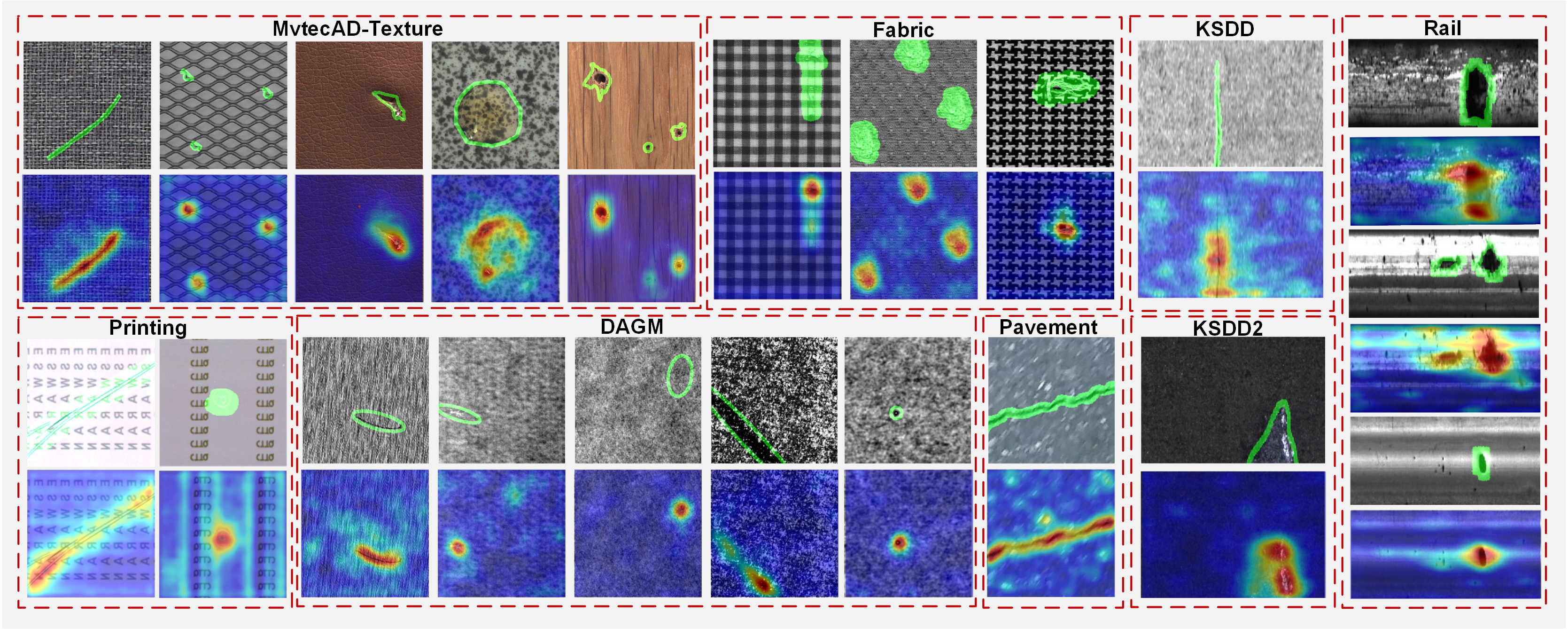}}
\caption[width=180mm]{
The qualitative detection outcomes of the proposed method on a variety of textured surfaces within the Texture Spectrum benchmark, the green border indicating the Ground Truth.
}
\label{fig2}
\end{figure*}

As depicted in Fig. 7(b) and (d), on the Mvtec-Texture dataset, the pixel-level AUROC of the model exhibits gradual improvement across all textures as $\eta$ increases. Conversely, on the Printing dataset, the pixel-level AUROC of patterns 1, 2, and 4 decreases as $\eta$ increases. This trend is attributed to the strengthened global distribution regularization with increasing $\eta$, resulting in an optimal transformation that yields reconstructed queries with more typical global normal characteristics. This enhancement is particularly beneficial for Mvtec-Texture datasets characterized by homogeneous textures, thereby improving model performance. However, for Printing data featuring anisotropic textures, there exists a significant disparity between its local features and global normal features. Consequently, increasing $\eta$ may lead to the failure of reconstructing normal local features, impacting performance negatively.

\subsection{Visualization}

\subsubsection{Visualization of regression behavior}
To provide a more intuitive validation of the proposed regression model, we performed T-SNE\cite{r51} visualization of its behavior on the MvtecAD-Texture dataset. The results are displayed in Fig. 8, presenting the feature distribution of the abnormal samples, normal samples, and their reconstruction results, as well as the extracted global support elements. Through this analysis, we made the following observations:
\begin{enumerate}
    \item The reconstructed features of normal samples closely resemble the original features. Conversely, the reconstructed features of abnormal samples deviate significantly from the original samples, showcasing notable differences.

    \item The feature distribution of normal samples aligns closely with the distribution of global support elements, indicating that these elements effectively capture the typical texture patterns.

    \item The reconstruction behavior of abnormal samples tends to align with the distribution of global support elements, demonstrating the guiding role of global normal regularization on the reconstruction of abnormal features.
\end{enumerate}

\subsubsection{Visualization of inspection results}

We illustrate the qualitative detection outcomes of our proposed method across different texture categories on the Texture Spectrum dataset in Fig. 9. The results demonstrate the accurate defect detection capability of our method on various types of textured surfaces, including both homogeneous and anisotropic textures, without the need for training.

\section{Conclusion}
In this paper, we introduce GRNR, a novel zero-shot learning approach for texture anomaly defect detection. Our method stands out for its data and labor efficiency, requiring no prior information such as training data or prompt to directly detect texture defects. The foundation of our approach lies in extracting intrinsic priors from images, specifically the global normal priors and local neighborhood priors. Leveraging these intrinsic supports, we propose the global-regularized neighborhood regression model and derive its efficient closed-form solution. To evaluate the efficacy of GRNR as an open-set zero-shot texture anomaly detection method for various textured surfaces, we constructed a comprehensive texture surface defect dataset called Texture Spectrum and conducted extensive experiments on it. Looking forward, our future work will delve into applying the zero-shot learning method for anomaly detection on objects.

\bibliographystyle{ieeetr} 
\bibliography{reference}

\vfill

\end{document}